\documentclass{article} 
\usepackage{iclr2025_conference,times}

\usepackage{amsmath,amsfonts,bm}

\def\eqref#1{equation~\ref{#1}}

\def\1{\bm{1}}

\DeclareMathAlphabet{\mathsfit}{\encodingdefault}{\sfdefault}{m}{sl}
\SetMathAlphabet{\mathsfit}{bold}{\encodingdefault}{\sfdefault}{bx}{n}

\input{mysymbol.sty}

\usepackage{hyperref}
\usepackage{url}
\usepackage{amsmath}
\usepackage{amssymb}
\usepackage{mathtools}
\usepackage{amsthm}
\usepackage{bbm}
\usepackage{xspace}
\usepackage{graphicx}
\usepackage[dvipsnames,svgnames]{xcolor}

\title{Fair Feature Importance Scores via Feature Occlusion and Permutation}

\author{Camille Little\textsuperscript{1}\thanks{Corresponding author: clittle1@rice.edu} , Madeline Navarro\textsuperscript{1}, Santiago Segarra\textsuperscript{1}, \& Genevera I. Allen\textsuperscript{2}\\
\textsuperscript{1}Department of Electrical and Computer Engineering, Rice University\\
\textsuperscript{2}Department of Statistics, Columbia University\\
}

\def \perm {{ \rho_{\rm\scriptscriptstyle perm} }}
\def \occl {{ \rho_{\rm\scriptscriptstyle occl} }}

\iclrfinalcopy 
\begin{document}

\maketitle

\begin{abstract}
As machine learning models increasingly impact society, their opaque nature poses challenges to trust and accountability, particularly in fairness contexts. Understanding how individual features influence model outcomes is crucial for building interpretable and equitable models. While feature importance metrics for accuracy are well-established, methods for assessing feature contributions to fairness remain underexplored. We propose two model-agnostic approaches to measure fair feature importance. First, we propose to compare model fairness before and after permuting feature values. This simple intervention-based approach decouples a feature and model predictions to measure its contribution to training. Second, we evaluate the fairness of models trained with and without a given feature. This occlusion-based score enjoys dramatic computational simplification via minipatch learning. Our empirical results reflect the simplicity and effectiveness of our proposed metrics for multiple predictive tasks. Both methods offer simple, scalable, and interpretable solutions to quantify the influence of features on fairness, providing new tools for responsible machine learning development.
\end{abstract}

\section{Introduction}
\label{gen_inst}

Many machine learning models deployed in society are often described as black boxes, lacking clear explanations for their decisions \citep{Adadi:2018}.
This problem is critical in the context of fairness, where lack of interpretability renders a model untrustworthy even if it appears to mitigate biased treatment.
We address this through one of the most common forms of interpretation: feature importance.
Our goal is to develop a model-agnostic feature importance score that captures how a single feature contributes to the fairness of a model's predictions.

Identifying important features is fundamental for efficient and interpretable machine learning.
Indeed, beyond aiding performance, feature importance can intuitively explain model behavior~\citep{Ribeiro:2016}.
Simpler tools, such as linear models, inherently quantify feature relevance, but the most useful methods may not provide obvious explanations.
Thus, many have proposed model-agnostic feature importance scores to measure how inputs affect predictions~\citep{Theng:2024}.
Some apply statistical tests to reliably identify relationships between features and predictions, albeit without giving the strength of their effects~\citep{Tansey:2022,Candes:2018}.
A more practical method for machine learning is directly measuring differences in outcomes when features are altered~\citep{Datta:2016,Datta:2017,Fisher:2019}.
Relevant approaches include importance scores based on permutations~\citep{Altmann:2010,Brieman:2001,Datta:2016}, Shapley values~\citep{Lundberg:2018}, and occlusion~\citep{FENG:2013,Lei:2018}.
A thorough overview on similarly defined measurements was compiled by~\cite{Covert:2021}.

In addition to affecting performance, features correlated with sensitive attributes can contribute to model bias, even if sensitive information is unused~\citep{Datta:2017, Zhao:2022}.
While determining important features is well-established for performance, measuring their influence on model bias is rare.
Recent metrics naturally extend existing feature importance scores, such as intuitive but computationally expensive Shapley values~\citep{Begley:2020} or interpretable but model-specific tree-based metrics~\citep{Little:2024}.
In a similar vein, fair feature selection searches for relevant subsets of features while preserving unbiased behavior~\citep{GrgicHlavca:2018,Galhotra:2017,Belitz:2021,Ling:2024,Brookhouse:2023,Bhargava:2020}, but they do not explicitly provide feature importance scores.

We propose two simple approaches to measure feature influence on model fairness. 
We consider (1) a \emph{permutation-based} score by comparing model fairness before and after randomly perturbing a given feature's values, and (2) an \emph{occlusion-based} score, where we compare the fairness of models trained with and without a given feature.
Both metrics are simple ways to approximately decouple features and model predictions.
Indeed, permutation and occlusion are intuitive yet well-founded techniques for feature importance~\citep{Datta:2016,Candes:2018,Fisher:2019}.
Moreover, our proposed scores yield different advantages: permutation retains the perturbed feature while preserving its marginal distribution, but occlusion is amenable to data splitting techniques, including the efficient approach of \emph{minipatch learning}~\citep{yao:2020}.
Finally, both metrics are not only suitable for any fairness metric and machine learning model, but they are also flexible in task, that is, we may apply them for time-series predictions, recommendations, or many other problems.

\section{Fair Feature Importance Scores}

Given data matrix $\bbX \in \reals^{N \times M} = [\bbX_1,\dots,\bbX_M]$ of $N$ samples with $M$ features, we wish to predict an unknown response vector $\bby\in\ccalY^N$ through a learned mapping $f:\reals^M\rightarrow \ccalY$, where $\ccalY$ dictates the task of interest, say as a discrete set of classes or real-valued regression targets.
Then, each sample is also associated with a sensitive attribute $z_i \in \{ 1,2,\dots,G \}$, representing the group to which the $i$-th sample belongs.
A model $f$ is considered biased if its predictions $\hat{\bby} = f(\bbX)$ exhibit dependence on the sensitive attribute $\bbz$~\citep{Kearns:2018}.
However, detecting bias through metrics such as demographic parity~\citep{Feldman:2015} does not indicate which features contributed to unfair predictions.
Hence, we propose two methods to evaluate feature influence on model bias.

\subsection{Importance Scores via Permutation}
\label{sec:perm}

We first define our permutation-based fair feature importance score as follows.

\textbf{Definition 1.} Let \(\bbX_{\pi(j)}\) be \(\mathbf{X}\) with the values of feature \(j\) randomly permuted across observations.
For a given bias metric $h$, the permutation-based importance score of feature $j$ is
\[
\perm(j) = 
h(\mathbf{y}, f_{\pi(j)}(\mathbf{X}_{\pi(j)}), \mathbf{z})
-
h(\mathbf{y}, f(\mathbf{X}), \mathbf{z}),
\]
where $f_{\pi(j)}$ and $f$ denote models trained on $\bbX_{\pi(j)}$ and $\bbX$, respectively.

This score measures the effect on fairness due to correlations with feature $j$ since the permutation decouples the feature $\bbX_j$ and the target $\bby$ while preserving the marginal distributions of all features.
Observe that $\perm(j)$ follows the intuition from~\citet{Romano:2020,Datta:2016,Fisher:2019}, where a given feature is sampled from an alternative distribution and the difference in a chosen metric is evaluated.
\cite{Fisher:2019} and \cite{Datta:2016} similarly consider permutations for feature influence, although they do not compare separately trained models.
Moreover, neither computes differences in bias upon permuting a given feature.


Thus, our metric $\perm$ can evaluate the extent to which feature $j$ influences model fairness.
While $\perm$ is simple and intuitive, computing such a metric for all features requires training $M+1$ different models leaving one feature out at a time, which may be computationally prohibitive for high-dimensional data.
In addition, $\perm$ may not reflect the contribution of $\bbX_j$ if it is correlated with other features along with $\bbz$.
If these correlations are known or inferred, we may permute multiple features simultaneously~\citep{Datta:2016,Fisher:2019}.
Instead, we consider an efficient, automated alternative via our second proposed metric.

\subsection{Importance Scores via Occlusion}
\label{sec:occlusion}  

Next, we define occlusion-based fair feature importance.

\textbf{Definition 2.} Let \(\bbX_{-j}\) denote \(\mathbf{X}\) without the $j$-th feature.
For a given bias metric $h$, the occlusion-based fair feature importance score of feature $j$ is
\[
\occl(j) = 
h(\mathbf{y}, f_{-j}(\mathbf{X}_{-j}), \mathbf{z})
-
h(\mathbf{y}, f(\mathbf{X}), \mathbf{z}),
\]
where $f_{-j}$ and $f$ denote models trained on $\bbX_{-j}$ and $\bbX$, respectively.

Our metric $\occl$ aligns with other leave-one-out-based scores~\citep{Lei:2018,Gevrey:2003}, which are well-understood for model performance but underused for fairness.
Moreover, $\perm$ requires sufficiently many samples $N$ to adequately decouple features and predictions, whereas $\occl$ is able to ignore a given feature even under low-sample regimes.


Thus, we propose to dramatically simplify computing $\occl$ using minipatch learning \citep{yao:2020}.
In particular, we sample $K$ minipatches, or submatrices $\bbX^{(k)} \in \reals^{n\times m}$ of $\bbX$ for $n < N$ and $m < M$, where $F_k \subseteq \{1,2,\dots,M\}$ denotes the row indices of $\bbX$ from which $\bbX^{(k)}$ is sampled.
For each minipatch $\bbX^{(k)}$, we train a model and evaluate the bias metric $h$ using all samples not in $F_k$, with $\hat{b}^{(k)}$ representing the obtained bias.
We then evaluate the fairness of the ensemble $\hat{b} = \frac{1}{K} \sum_{k=1}^K \hat{b}^{(k)}$.
For the fairness upon occlusion of feature $j$, we compute
\begin{align*}
    \hat{b}_{-j} = \frac{1}{\sum_{k=1}^{K} \mbI ( j \notin F_k ) } \sum_{k=1}^{K} \mbI (j \notin F_k ) \hat{b}^{(k)},
\end{align*}
where $\mbI(\cdot)$ represents the indicator function taking value 1 when its argument is true and 0 otherwise.
Thus, $\hat{b}_{-j}$ computes the fairness across all minipatch models except those containing feature $j$.
Then, we compute the minipatch-based score $\occl$ as $\hat{b}_{-j} - \hat{b}$.
This gives us an efficient way to compute the fair occlusion feature importance score with no additional computation required. 
Note that we can analogously compute a feature importance score for accuracy as well.


\begin{figure}[t!]
    \centering
    \includegraphics[width =\textwidth]{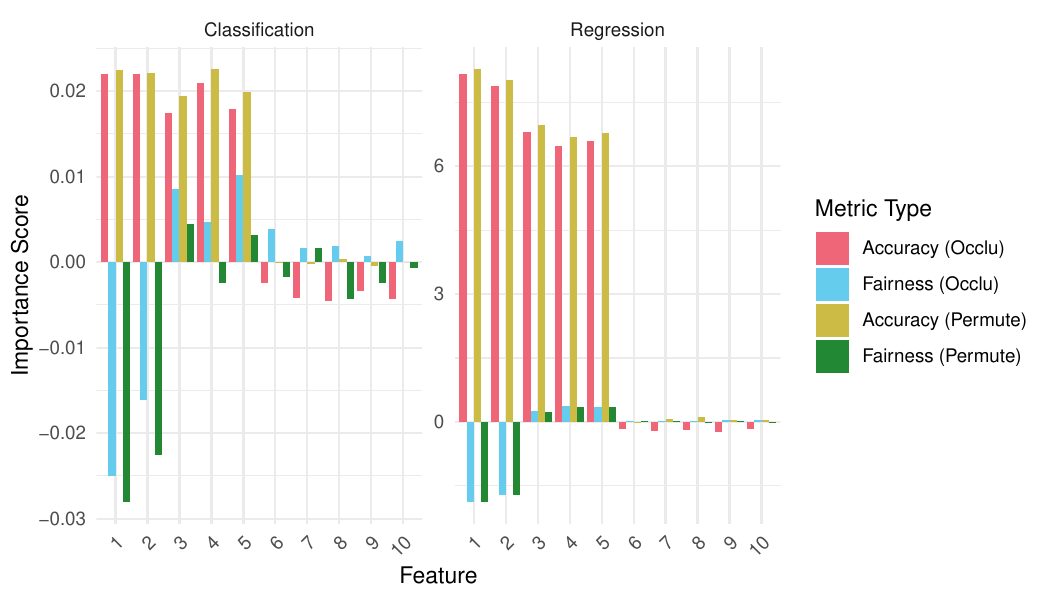}
    \caption{
    Classification and regression results are presented for accuracy, measured by classification error and MSE, and fairness, measured by demographic parity~\citep{Feldman:2015}, using a Random Forest model. Feature importance scores were computed using occlusion $\occl$ and permutation $\perm$ metrics for both simulation types. In these simulations, the first five features are signal features associated with the outcome, while the first two are correlated with the protected attribute, introducing bias. Positive scores indicate that a feature improves fairness or accuracy, while negative scores suggest the opposite. The magnitudes and directions of the scores align as expected, consistent with the simulation design.}
    \label{fig:sims}
\end{figure}

\section{Empirical Studies}


\subsection{Simulation setup and Results}

We develop a synthetic data model to test the validity of our fair feature importance score for both classification and regression tasks.
We consider data with $N=1,\!000$ samples and $M=10$ features, where the first five features are semantically relevant and the first two are correlated with the sensitive attribute $\bbz$.
More specifically, we first generate binary protected attributes $z_i \sim \text{Bernoulli}(0.2)$ for every sample.
Then, for every entry of $\bbX$, we simulate $X_{ij} \sim \ccalN( (1 + z_i)\, \mbI(j\leq 2), 1 )$, that is, $\bbX_j$ contains mean shifts correlated with $\bbz$ for $j \leq 2$, otherwise $\bbX_j$ denotes zero-mean white noise for $j > 2$.
We generate classification labels via a logistic regression model, where $\bby = \text{logit}(\bbX\bbbeta)$ for $\beta_j \sim \ccalN(5 \, \mbI( j\leq 5 ), 0.1 )$.
Thus, the signal features $\bbX_j$ for $j \leq 5$ are correlated with $\bby$ while the remainder are irrelevant.
For regression, we apply a linear model with Gaussian noise, that is, $\bby = \bbX\bbbeta + \bbepsilon$, where $\epsilon_i \sim \ccalN(0,1)$.

Figure \ref{fig:sims} presents the accuracy and fairness scores computed via permutation and occlusion methods using a Random Forest model for both classification and regression tasks. Negative fairness scores indicate features that harm fairness, while positive scores suggest features that improve it. The accuracy scores follow a similar interpretation, with higher values indicating greater importance for prediction. In the simulation, the first two features are correlated with the protected attribute, introducing bias. As expected, both occlusion (blue) and permutation (green) fairness metrics yield strongly negative values for these features. The first five features are signal features, essential for predicting the outcome, as reflected by their positive accuracy scores for both occlusion (red) and permutation (yellow).  It is important to note that the metrics for regression and classification are on different scales due to their distinct objective functions. Classification scores tend to exhibit greater variability, likely due to the hard decision boundaries inherent in classification models, which can amplify sensitivity to small changes in feature values.
\begin{figure}[t!]
    \centering
    \includegraphics[width =\textwidth]{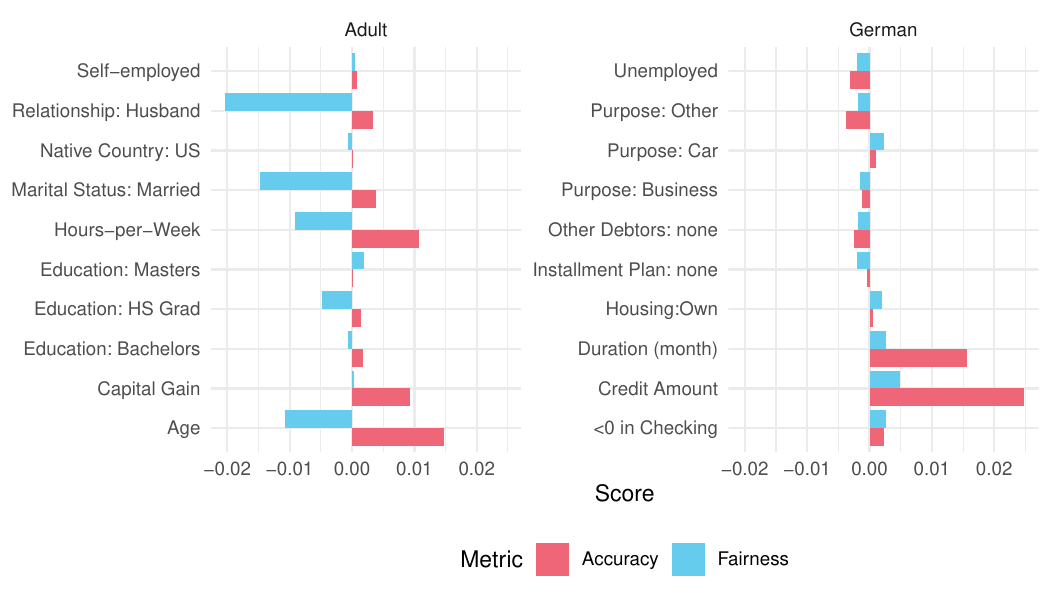}
    \caption{Random Forest interpretation using occlusion scores via minipatches for the Adult Income Dataset and German Credit dataset. The importance scores' magnitudes and directions align with other studies done on these datasets.}
    \label{fig:case_study}
\end{figure}

\subsection{Case Studies}
To align our work with the existing fairness literature, we evaluate our occlusion-based metric on two popular UCI fairness benchmark datasets \citep{Dua:2017}. We examine: (i) the Adult Income dataset containing 96 one-hot-encoded features and  45,222 individuals with class labels stating whether their income is greater than \$50,000 and Gender as the protected attribute; and (ii) the German Credit dataset, which classifies people with good or bad credit risks based on 56 one-hot-encoded features and 1,000 observations and we use Gender as the protected attribute. Given the large number of features in these datasets, we leverage minipatch learning to compute our scores efficiently. 
This is similar in concept to the Shapley-based approach in~\citep{Begley:2020}, but with minipatches, $\occl$ is more efficient.
We do not include fair Shapley feature importance in our simulations as it can be computationally prohibitive for high-dimensional datasets, and their code is not publicly available.
We set the minipatch size to $n = 0.2N$ and $m = 0.2M$, using $K = 2,\!000$ minipatches. Our base learner for the experiments is a Random Forest classifier.

The left-hand side of Figure \ref{fig:case_study} shows fairness and accuracy importance scores for the Adult Income dataset, while the right-hand side presents the scores for the German Credit dataset. 
In the Adult Income dataset, the ``Relationship: Husband" feature has the most negative fairness score, indicating that it is the most biased feature. 
This is very natural, since it is directly related to gender, the protected attribute.
Conversely, ``Capital Gain'' emerges as one of the most predictive features, reflecting that financial indicators often carry strong signals for income prediction.  Interestingly, the ``hours-per-week" feature demonstrates a fairness-accuracy tradeoff, being both biased and highly important for accuracy. This tradeoff highlights the complexity of balancing predictive performance with fairness, as labor market factors may disproportionately affect specific demographic groups. The random forest model for the Adult dataset achieves an accuracy of 0.84 and a fairness score of 0.85, suggesting that while there is potential to enhance fairness by removing biased features, doing so will likely come at the expense of predictive accuracy.   

In contrast, the German Credit dataset paints a different picture. Features such as ``Duration of the Loan" and ``Credit Amount" are both fair and predictive for accuracy. This outcome aligns with the model's relatively high fairness score of 0.94 alongside an accuracy of 0.8, indicating that the dataset may inherently present less conflict between fairness and predictive power. The absence of a pronounced fairness-accuracy tradeoff suggests that financial attributes in credit risk assessment are more equitably distributed across demographic groups compared to factors influencing income prediction.  

\section{Discussion and Future Work}

We proposed two simple and intuitive model-agnostic metrics for quantifying feature contribution to model bias.
Our feature importance scores complement other popular interpretable feature contribution measurements as they are effective, efficient, and easy to understand.
In our synthetic simulations, our metrics properly decrease when features are correlated with sensitive attributes.
Furthermore, our real-data analysis allowed us to determine features most related to gender bias.
We also found that our proposed metrics can identify relevant, unbiased features for real-world data, implying cases where feature selection may improve both fairness and accuracy.
Future work will see extensions to measuring the bias contributions of multiple features, including evaluating when interactions of features affect fairness \citep{little:2025}. 

\subsubsection*{Acknowledgments}
The authors acknowledge support from NSF DMS-2210837 and the JP Morgan Faculty Research Award. CL acknowledges support from the NSF Graduate Research Fellowship Program under grant number 1842494. This work was partially supported by NSF under award CCF-2340481. Research was sponsored by the Army Research Office and was accomplished under Grant Number W911NF-17-S-0002. The views and conclusions contained in this document are those of the authors and should not be interpreted as representing the official policies, either expressed or implied, of the Army Research Office or the U.S. Army or the U.S. Government. The U.S. Government is authorized to reproduce and distribute reprints for Government purposes notwithstanding any copyright notation herein.
\newpage
\bibliography{iclr2025_conference}

\begin{thebibliography}{30}
\providecommand{\natexlab}[1]{#1}
\providecommand{\url}[1]{\texttt{#1}}
\expandafter\ifx\csname urlstyle\endcsname\relax
  \providecommand{\doi}[1]{doi: #1}\else
  \providecommand{\doi}{doi: \begingroup \urlstyle{rm}\Url}\fi

\bibitem[Adadi \& Berrada(2018)Adadi and Berrada]{Adadi:2018}
Amina Adadi and Mohammed Berrada.
\newblock Peeking inside the black-box: A survey on explainable artificial intelligence ({XAI}).
\newblock \emph{IEEE Access}, 6:\penalty0 52138--52160, 2018.

\bibitem[Altmann et~al.(2010)Altmann, Toloşi, Sander, and Lengauer]{Altmann:2010}
André Altmann, Laura Toloşi, Oliver Sander, and Thomas Lengauer.
\newblock Permutation importance: A corrected feature importance measure.
\newblock \emph{Bioinformatics}, 26:\penalty0 1340--1347, 2010.

\bibitem[Begley et~al.(2020)Begley, Schwedes, Frye, and Feige]{Begley:2020}
Tom Begley, Tobias Schwedes, Christopher Frye, and Ilya Feige.
\newblock Explainability for fair machine learning.
\newblock \emph{ArXiv Pre-Print 2010.07389}, 2020.

\bibitem[Belitz et~al.(2021)Belitz, Jiang, and Bosch]{Belitz:2021}
Clara Belitz, Lan Jiang, and Nigel Bosch.
\newblock Automating procedurally fair feature selection in machine learning.
\newblock In \emph{AAAI/ACM Conference on AI, Ethics, and Society}, pp.\  379--389, 2021.

\bibitem[Bhargava et~al.(2020)Bhargava, Couceiro, and Napoli]{Bhargava:2020}
Vaishnavi Bhargava, Miguel Couceiro, and Amedeo Napoli.
\newblock {{LimeOut}}: An ensemble approach to improve process fairness.
\newblock In \emph{ECML PKDD 2020 Workshops}, pp.\  475--491, 2020.

\bibitem[Breiman(2001)]{Brieman:2001}
Leo Breiman.
\newblock Random forests.
\newblock \emph{Machine Learning}, 45\penalty0 (1):\penalty0 5--32, 2001.

\bibitem[Brookhouse \& Freitas(2023)Brookhouse and Freitas]{Brookhouse:2023}
James Brookhouse and Alex Freitas.
\newblock Fair feature selection: A comparison of multi-objective genetic algorithms.
\newblock \emph{ArXiv Pre-Print 2310.02752}, 2023.

\bibitem[Candès et~al.(2018)Candès, Fan, Janson, and Lv]{Candes:2018}
Emmanuel Candès, Yingying Fan, Lucas Janson, and Jinchi Lv.
\newblock Panning for gold: {‘Model-X’} knockoffs for high dimensional controlled variable selection.
\newblock \emph{Journal of the Royal Statistical Society Series B: Statistical Methodology}, 80\penalty0 (3):\penalty0 551--577, 01 2018.

\bibitem[Covert et~al.(2021)Covert, Lundberg, and Lee]{Covert:2021}
Ian Covert, Scott Lundberg, and Su-In Lee.
\newblock Explaining by removing: A unified framework for model explanation.
\newblock \emph{Journal of Machine Learning Research (JMLR)}, 22\penalty0 (209):\penalty0 1--90, 2021.

\bibitem[Datta et~al.(2016)Datta, Sen, and Zick]{Datta:2016}
Anupam Datta, Shayak Sen, and Yair Zick.
\newblock Algorithmic transparency via quantitative input influence: Theory and experiments with learning systems.
\newblock In \emph{IEEE Symposium on Security and Privacy (SP)}, pp.\  598--617, 2016.

\bibitem[Datta et~al.(2017)Datta, Fredrikson, Ko, Mardziel, and Sen]{Datta:2017}
Anupam Datta, Matt Fredrikson, Gihyuk Ko, Piotr Mardziel, and Shayak Sen.
\newblock Proxy non-discrimination in data-driven systems.
\newblock \emph{ArXiv Pre-Print 1707.08120}, 2017.

\bibitem[Dua \& Graff(2017)Dua and Graff]{Dua:2017}
Dheeru Dua and Casey Graff.
\newblock {UCI} machine learning repository, 2017.
\newblock URL \url{http://archive.ics.uci.edu/ml}.

\bibitem[Feldman et~al.(2015)Feldman, Friedler, Moeller, Scheidegger, and Venkatasubramanian]{Feldman:2015}
Michael Feldman, Sorelle~A. Friedler, John Moeller, Carlos Scheidegger, and Suresh Venkatasubramanian.
\newblock Certifying and removing disparate impact.
\newblock In \emph{ACM SIGKDD International Conference on Knowledge Discovery and Data Mining}, pp.\  259--268, 2015.

\bibitem[Feng et~al.(2013)Feng, Chen, and Xu]{FENG:2013}
Dingcheng Feng, Feng Chen, and Wenli Xu.
\newblock Efficient leave-one-out strategy for supervised feature selection.
\newblock \emph{Tsinghua Science and Technology}, 18:\penalty0 629--635, 2013.

\bibitem[Fisher et~al.(2019)Fisher, Rudin, and Dominici]{Fisher:2019}
Aaron Fisher, Cynthia Rudin, and Francesca Dominici.
\newblock All models are wrong, but many are useful: Learning a variable's importance by studying an entire class of prediction models simultaneously.
\newblock \emph{Journal of Machine Learning Research (JMLR)}, 20\penalty0 (177):\penalty0 1--81, 2019.

\bibitem[Galhotra et~al.(2017)Galhotra, Brun, and Meliou]{Galhotra:2017}
Sainyam Galhotra, Yuriy Brun, and Alexandra Meliou.
\newblock Fairness testing: Testing software for discrimination.
\newblock In \emph{Joint Meeting on Foundations of Software Engineering (ESEC/FSE)}, pp.\  498––510, 2017.

\bibitem[Gevrey et~al.(2003)Gevrey, Dimopoulos, and Lek]{Gevrey:2003}
Muriel Gevrey, Ioannis Dimopoulos, and Sovan Lek.
\newblock Review and comparison of methods to study the contribution of variables in artificial neural network models.
\newblock \emph{Ecological Modelling}, 160\penalty0 (3):\penalty0 249--264, 2003.

\bibitem[{Grgi{\'c}-Hla{\v c}a} et~al.(2018){Grgi{\'c}-Hla{\v c}a}, Zafar, Gummadi, and Weller]{GrgicHlavca:2018}
Nina {Grgi{\'c}-Hla{\v c}a}, Muhammad~Bilal Zafar, Krishna~P. Gummadi, and Adrian Weller.
\newblock Beyond distributive fairness in algorithmic decision making: Feature selection for procedurally fair learning.
\newblock \emph{AAAI Conference on Artificial Intelligence}, 32\penalty0 (1), 2018.

\bibitem[Kearns et~al.(2018)Kearns, Neel, Roth, and Wu]{Kearns:2018}
Michael Kearns, Seth Neel, Aaron Roth, and Zhiwei~Steven Wu.
\newblock Preventing fairness gerrymandering: Auditing and learning for subgroup fairness.
\newblock In \emph{International Conference on Machine Learning (ICML)}, volume~80, pp.\  2564--2572, 2018.

\bibitem[Lei et~al.(2018)Lei, G’Sell, Rinaldo, Tibshirani, and Wasserman]{Lei:2018}
Jing Lei, Max G’Sell, Alessandro Rinaldo, Ryan~J. Tibshirani, and Larry Wasserman.
\newblock Distribution-free predictive inference for regression.
\newblock \emph{Journal of the American Statistical Association}, 113\penalty0 (523):\penalty0 1094--1111, 2018.

\bibitem[Ling et~al.(2024)Ling, Xu, Zhou, Du, Yu, and Wu]{Ling:2024}
Zhaolong Ling, Enqi Xu, Peng Zhou, Liang Du, Kui Yu, and Xindong Wu.
\newblock Fair feature selection: A causal perspective.
\newblock \emph{ACM Transactions on Knowledge Discovery from Data}, 18\penalty0 (7):\penalty0 1--23, 2024.

\bibitem[Little et~al.(2025)Little, Zheng, and Allen]{little:2025}
Camille Little, Lili Zheng, and Genevera Allen.
\newblock {iLOCO}: Distribution-free inference for feature interactions.
\newblock \emph{ArXiv Pre-Print 2502.06661}, 2025.

\bibitem[Little et~al.(2024)Little, Lina, and Allen]{Little:2024}
Camille~Olivia Little, Debolina~Halder Lina, and Genevera~I. Allen.
\newblock Fair feature importance scores for interpreting decision trees.
\newblock \emph{Transactions on Machine Learning Research}, 2024.

\bibitem[Lundberg et~al.(2018)Lundberg, Erion, and Lee]{Lundberg:2018}
Scott~M. Lundberg, Gabriel~G. Erion, and Su{-}In Lee.
\newblock Consistent individualized feature attribution for tree ensembles.
\newblock \emph{ArXiv Pre-Print 1802.03888}, 2018.

\bibitem[Ribeiro et~al.(2016)Ribeiro, Singh, and Guestrin]{Ribeiro:2016}
Marco~Tulio Ribeiro, Sameer Singh, and Carlos Guestrin.
\newblock ``{Why} should {I} trust you?'': Explaining the predictions of any classifier.
\newblock In \emph{ACM SIGKDD International Conference on Knowledge Discovery and Data Mining}, pp.\  1135–1144, 2016.

\bibitem[Romano et~al.(2020)Romano, Bates, and Candes]{Romano:2020}
Yaniv Romano, Stephen Bates, and Emmanuel Candes.
\newblock Achieving equalized odds by resampling sensitive attributes.
\newblock In \emph{Advances in Neural Information Processing Systems (NeurIPS)}, volume~33, pp.\  361--371, 2020.

\bibitem[Tansey et~al.(2022)Tansey, Veitch, Zhang, Rabadan, and Blei]{Tansey:2022}
Wesley Tansey, Victor Veitch, Haoran Zhang, Raul Rabadan, and David~M. Blei.
\newblock The holdout randomization test for feature selection in black box models.
\newblock \emph{Journal of Computational and Graphical Statistics}, 31\penalty0 (1):\penalty0 151--162, 2022.

\bibitem[Theng \& Bhoyar(2024)Theng and Bhoyar]{Theng:2024}
Dipti Theng and Kishor~K. Bhoyar.
\newblock Feature selection techniques for machine learning: A survey of more than two decades of research.
\newblock \emph{Knowledge and Information Systems}, 66\penalty0 (3):\penalty0 1575--1637, 2024.

\bibitem[Yao \& Allen(2020)Yao and Allen]{yao:2020}
Tianyi Yao and Genevera~I. Allen.
\newblock Feature selection for huge data via minipatch learning.
\newblock \emph{ArXiv Pre-Print 2010.08529}, 2020.

\bibitem[Zhao et~al.(2022)Zhao, Dai, Shu, and Wang]{Zhao:2022}
Tianxiang Zhao, Enyan Dai, Kai Shu, and Suhang Wang.
\newblock Towards fair classifiers without sensitive attributes: Exploring biases in related features.
\newblock In \emph{ACM International Conference on Web Search and Data Mining}, pp.\  1433––1442, 2022.

\end{thebibliography}
\bibliographystyle{iclr2025_conference}


\end{document}